\documentclass{article}

\usepackage[nonatbib,final]{nips_2017}

\usepackage[utf8]{inputenc} % allow utf-8 input
\usepackage[T1]{fontenc}    % use 8-bit T1 fonts
\usepackage{hyperref}       % hyperlinks
\usepackage{url}            % simple URL typesetting
\usepackage{booktabs}       % professional-quality tables
\usepackage{bm}
\usepackage{bbm}
\usepackage{amsfonts}       % blackboard math symbols
\usepackage{nicefrac}       % compact symbols for 1/2, etc.
\usepackage{microtype}      % microtypography
\usepackage{amsmath}
\usepackage{amssymb}
\usepackage{amsfonts}
\usepackage{algorithm,algorithmic}
\usepackage{graphicx}
\usepackage{caption}
\usepackage{subcaption}

\title{Mondrian Processes for Flow Cytometry Analysis}

\author{Disi Ji, Eric Nalisnick, Padhraic Smyth \\
\small{Department of Computer Science}\\
\small{University of California, Irvine}\\
\texttt{\{disij, enalisni, p.smyth\}@uci.edu} }

\begin{document}

\maketitle

 \begin{abstract}
 Analysis of flow cytometry data is an essential tool for clinical diagnosis of hematological and immunological conditions.  Current clinical workflows rely on a manual process called \textit{gating} to classify cells into their canonical types.  This dependence on human annotation limits the rate, reproducibility, and complexity of flow cytometry analysis.  In this paper, we propose using Mondrian processes to perform automated gating by incorporating prior information of the kind used by gating technicians. The method segments cells into types via Bayesian nonparametric trees. Examining the posterior over trees allows for interpretable visualizations and uncertainty quantification---two vital qualities for implementation in clinical practice.
\end{abstract}

\vspace{-0.2in}
\section{Introduction}
\vspace{-0.1in}
Blood cancers such as leukemia and lymphoma \cite{wu2013flow}, and immunodeficiencies such as HIV infection \cite{abraham2016flow}, are routinely diagnosed with the help of \textit{flow cytometry} (FC).  FC provides high dimensional feature representations of blood cells based on their light scattering and fluorescence properties \cite{o2013flow}.  These features can then be used to diagnose blood conditions by identifying cell sub-populations of a canonical type (such as T-cell variants). These types are identified by \textit{manual gating} \cite{verschoor2015introduction}: technicians visually inspect two-dimensional scatter plots of the data and draw  bounding boxes around potential clusters.  The cells that fall within a box are then plotted again in another scatter plot involving two dimensions different from those previously inspected, and so on in a recursive manner.  
%The box-drawing process continues in a recursive manner stopping when the final scatter plot isolates a specific sub-population that can inform a diagnosis.  
%Unfortunately, due to space constraints, we refer the reader to O'Neill et al.\ (2013) \cite{o2013flow} for an introduction to flow cytometry.
%While FC is an essential tool for hematology and immunology, 

Manual gating has limitations in terms of its effectiveness and efficiency  \cite{aghaeepour2013critical,verschoor2015introduction,saeys2016computational}. One significant drawback is that the method is subjective and heuristic in nature, and hence, is not reproducible. In addition, even for expert gaters, as FC technology scales to 50+ features,
%the combinatorics of searching for subpopulations via visual inspection of subspaces quickly becomes unmanageable and 
human analysis becomes significantly slower  and less likely to take full advantage of the enriched measurement space \cite{chester2015algorithmic}. 
%Regarding the former, examining only two dimensions at a time ignores any three or more dimensional correlations in the features.  Regarding the latter, the work-flow's dependence on human specialists bottlenecks the rate of analysis (as the number of trained technicians is limited) as well as makes the process unable to be rigorously standardized.  In the future, as an increasing number of cell-level features are extracted by flow cytometers ($100+$ dimensions), human analysis will only become slower and less likely to take full advantage of the enriched measurement space. 
Replacing or augmenting gating with a machine learning algorithm is  appealing,  as an algorithm can in principle provide a more reproducible, efficient, and thorough analysis of the data \cite{aghaeepour2013critical, qiu2011extracting}.  However, 
existing automated approaches also have drawbacks.
%due to FC's crucial role in clinical practice, making such a drastic change to the diagnosis process is difficult.  The foremost impediment is the black-box nature of most machine learning algorithms: a clinician likely would not (and probably should not) trust the algorithm's prediction if he or she cannot inspect the algorithm's decision-making process. 
For example, existing clustering and dimensionality reduction methods for FC analysis (e.g., \cite{pyne2009automated, o2013flow,aghaeepour2013critical,amir2013visne, naim2014swift,mair2016end}) produce results that are difficult to interpret in comparison to the simplicity of gating.  In addition, existing approaches are, in general, unable to incorporate the prior knowledge that gating technicians use to help decide where to draw the bounding boxes and which pair of dimensions to plot next \cite{qiu2011extracting}.  For instance, a technician, through his/her training, would know that B-cells exhibit high values for marker CD19 and low values for marker CD3 \cite{lee2017automated}.  

In this paper, we propose using \textit{Mondrian processes} (MPs) \cite{roy2009mondrian} for classifying cells into a canonical type.  MPs define a Bayesian nonparametric prior over kd-trees and represent a useful bridge between manual gating and machine learning.  MPs are interpretable as their generative process closely follows the gating procedure: MPs partition a space by sampling a dimension and then drawing an axis-aligned cut through that dimension.  Moreover, as the MP is a Bayesian model, existing knowledge can be incorporated as prior distributions.  As we explain in Section \ref{model}, we incorporate expert knowledge into the distribution over where to draw the cut.  Performing posterior inference results in a distribution over cuts/partitions, and thus, model uncertainty can be quantified and even visualized for clinicians, building their trust in the model.          

\begin{figure}
    \begin{subfigure}{.35\textwidth}
    \centering
    \includegraphics[width=.99\linewidth]{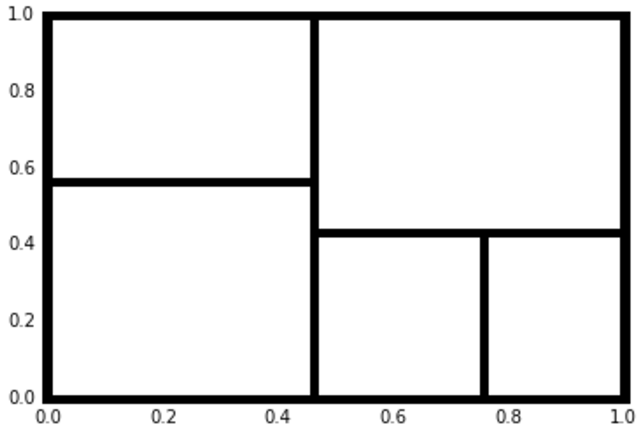}
    \caption{$\hat{\mathcal{M}} \sim \text{MP}(\lambda_{0}=1, \mathcal{X}=[0,1]\times[0,1])$}
    \end{subfigure}
    \hfil
    \begin{subfigure}{.5\textwidth}
    \begin{algorithm}[H]
    \begin{tiny}
    \begin{algorithmic}
    \STATE \textbf{Input}: Lifetime $\lambda_{0}$, space to partition $\mathcal{X}$, table of prior information $\mathcal{T}_{C \times D}$.  \\
    \texttt{\small{Sample\_Mondrian}}($\lambda_{0}$, $\mathcal{X}$, $\mathcal{T}_{C \times D}$):\\
    \hspace{1em} $\hat{t} \sim \text{Exponential}(\sum_{d} \gamma_{\mathcal{T}[d]} | \mathcal{X}_{d} |)$\\
    \hspace{1em} $\lambda \leftarrow \lambda - \hat{t}$\\
    \hspace{1em} \textbf{if} $\lambda < 0$ or $|\mathcal{T}| = 1$:\\
    \hspace{2em} \textbf{return} ($\mathcal{X}$, $\emptyset$, $\emptyset$)\\
    \hspace{1em} $\hat{d} \sim \text{Multinoulli}(p_{d} \propto \gamma_{\mathcal{T}[d]}| \mathcal{X}_{d} |)$\\
    \hspace{1em} $\hat{r} \sim \text{Beta}(\alpha=\phi_{\mathcal{T}[d]}, \beta=\phi_{\mathcal{T}[d]})$\\
    \hspace{1em} $\hat{c} \leftarrow a_{d} + \hat{r}(b_{d} - a_{d})$\\
    \hspace{1em} $\mathcal{X}_{<c} \leftarrow \{ \mathcal{X}_{1}\times \ldots \times \mathcal{X}_{\hat{d}}^{< \hat{c}} \times \ldots \times \mathcal{X}_{D} \}$\\
    \hspace{1em} $\mathcal{X}_{>c} \leftarrow \{ \mathcal{X}_{1}\times \ldots \times \mathcal{X}_{\hat{d}}^{> \hat{c}} \times \ldots \times \mathcal{X}_{D} \}$\\
    \hspace{1em} $\mathcal{T}_{d-} \leftarrow \mathcal{T}[$\texttt{select * where }$\mathcal{T}[\hat{d}]=-1$\texttt{ or }$0]$\\
    \hspace{1em} $\mathcal{T}_{d+} \leftarrow \mathcal{T}[$\texttt{select * where }$\mathcal{T}[\hat{d}]=+1$\texttt{ or }$0]$\\
    \hspace{1em} $\mathcal{M}_{<c} \leftarrow$ \texttt{\small{Sample\_Mondrian}}($\lambda$, \ $\mathcal{X}_{< c}$, $\mathcal{T}_{d-}$)\\
    \hspace{1em} $\mathcal{M}_{>c} \leftarrow$ \texttt{\small{Sample\_Mondrian}}($\lambda$, \ $\mathcal{X}_{> c}$, $\mathcal{T}_{d+}$)\\
    \hspace{1em} \textbf{return} ($\mathcal{X}$, $\mathcal{M}_{< c}$, $\mathcal{M}_{> c}$)
    \end{algorithmic}
    \caption{Mondrian Process with Flow Cytometry Priors}
    \end{tiny}
    \end{algorithm}
    \end{subfigure}

\caption{\textit{The Mondrian Processes}.  Subfigure (a) shows a draw from MP$(\lambda_{0}=1, \mathcal{X}=[0,1] \times [0,1])$} and subfigure (b) implements sampling from an MP with priors set by an information table $\mathcal{T}$.
\label{MPdef}
\end{figure}

\vspace{-0.1in}
\section{Mondrian Processes for Automated Gating}
\vspace{-0.1in}\label{model}
\noindent{\bf Background.}  We first review the relevant background material on Mondrian processes---see Balog and Teh (2015) \cite{balog2015mondrian} for a more thorough introduction.  The \textit{Mondrian process} (MP) \cite{roy2009mondrian} is a nonparametric process in which a finite region is segmented into rectangular partitions, resulting in a structure that looks like a painting by artist Piet Mondrian.  An example of a draw from a Mondrian process is shown in Figure \ref{MPdef} (a). 
%More formally, MPs are time-indexed stochastic processes that partition a product space via random axis-aligned cuts.  They can be thought of as a prior over $k$d-trees whose depth is determined by an exponential clock \cite{balog2016mondrian}: a time budget is proposed and cuts are made until time exceeds the budget---or in other words, the clock `rings'.  All cuts after the ring occurs are then ignored.  The box is then partitioned by the cuts into (hyper-) rectangles whose size is inversely proportional to the budget.  \textbf{Definition.}  

We now specify the processes formally.  Define an axis-aligned box $\mathcal{X}$ to be a product space of $D$ bounded intervals $\mathcal{X}_{d}=[a_{d}, b_{d}]$ with length $| \mathcal{X}_{d} | = b_{d} - a_{d}$: $\mathcal{X} = \{ \mathcal{X}_{1}\times \ldots \times \mathcal{X}_{D} \}$.  Define a \textit{Mondrian process} (MP) with a lifetime (budget) $\lambda_{0}$ and on a space $\mathcal{X}$ as $\text{MP}(\lambda_{0}, \mathcal{X})$.  The process proceeds by first drawing an exponential random variable $\hat{t} \sim \text{Exponential}(\sum_{d} |\mathcal{X}_{d}|)$.  If $\hat{t} > \lambda_{0}$, the process halts and returns $\mathcal{X}$ without any partitions.  If $\hat{t} < \lambda_{0}$, a dimension is drawn proportionally to its length ($p_{d} \propto |\mathcal{X}_{d}|$) and then a cut location $c$ is drawn according to $\hat{c} \sim \text{Uniform}([a_{d}, b_{d}])$.  In other words, the space is partitioned by $c$ into two new spaces $\mathcal{X}^{<c} = \{ \mathcal{X}_{1}\times \ldots \times \mathcal{X}_{\hat{d}}^{< \hat{c}} \times \ldots \times \mathcal{X}_{D} \}$ and  $\mathcal{X}^{>c} = \{ \mathcal{X}_{1}\times \ldots \times \mathcal{X}_{\hat{d}}^{> \hat{c}} \times \ldots \times \mathcal{X}_{D} \}$.  Two child processes $\text{MP}(\lambda-\hat{t}, \mathcal{X}^{<c})$ and $\text{MP}(\lambda-\hat{t}, \mathcal{X}^{>c})$ are then spawned and the process recurses with a decreased lifetime $\lambda' = \lambda_{0}-\hat{t}$ and on the subdomains $\mathcal{X}' = \mathcal{X}^{<c}$ and  $\mathcal{X}'' = \mathcal{X}^{>c}$.  This recursion arises from the elegant self-consistentency property of MPs: further cuts to any partition are themselves drawn from an MP with a lifetime and domain properly inherited from the parent process.  Below we descibe an extension of this standard MP, as motivated by the problem of cell classification.

%Clearly, due to the combinatorial nature of MPs, inference is challenging \cite{roy2009mondrian}.  In fact, the asymptotic identifiability of MPs is an open question \cite{roy2009mondrian}.  Previous work on modeling with MPs has either sampled from a fixed prior \cite{balog2016mondrian,lakshminarayanan2014mondrian} or used MCMC to collect posterior samples \cite{roy2009mondrian, wang2015metadata}. 
%\section{Mondrian Processes with Flow Cytometry Priors}\label{informed}
\noindent{\bf Representing Prior Information about Cell Types.}    Given a cell's coordinates in the multi-dimensional marker space, we wish to classify the cell into a canonical cell type.  Marker features are used because they are biologically motivated.  For example, CD4 T cells are named as such because they express the surface protein CD4 and thus exhibit a high response to the CD4 marker.  Unfortunately, the exact coordinates of a canonical cell type are not known precisely due to variation from measurement error and biological diversity.  Rather, the only general statement that can be made is that CD4 T cells will tend to have a high CD4 response when compared to other cells in the sample.  The prior information typically takes the form shown in subtable (a) of Figure \ref{priorInfo} \cite{lee2017automated}: $+1$ denotes a high expected response in a given marker (dimension), $-1$ a low expected response, and $0$ denotes no prior information.  

Our approach is motivated by the work of Lee et al.\ \cite{lee2017automated}, who recently proposed an algorithm, entitled \textit{ACDC}, for automated gating with prior information. The primary differences between our approach and ACDC are that (1) ACDC is semi-supervised, requiring   human-provided type labels for a set of landmark points, whereas our method does not require any cell-level labels, and (2) ACDC uses multiple algorithmic steps (unsupervised clustering, random walks on a nearest neighbors graph) in contrast to the more interpretable and coherent  generative model we propose here.

\noindent{\bf Generative Model: Mondrian Process with Gaussian Emissions.}  
We propose the following generative model for cell types in flow cytometry data.  Given a sample of N cells $\mathbf{X} = \{\mathbf{x}_{1},\ldots, \mathbf{x}_{N}\} (\mathbf{x}_{i} \in \mathbb{R}^{D})$ each with $D$ marker responses, we propose each cell be modeled as an i.i.d. draw from the following hierarchical process: \begin{equation}\begin{split}
    \{\mathcal{M}_{1},\ldots,\mathcal{M}_{K}\} &\sim \text{MP}(\mathcal{X}=[a_{1}, b_{1}] \times \ldots \times [a_{D}, b_{D}], \mathcal{T}_{C \times D}) \\ \mathbf{x}_{i} &\sim \sum_{k=1}^{K} \mathbbm{1}[\mathbf{x}_{i} \in \mathcal{M}_{k}] \  \text{Normal}(\boldsymbol{\mu}_{k}, \boldsymbol{\Sigma}_{k})
\end{split}\end{equation} where $\mathcal{X}$ is the observed ranges of the sample $\mathbf{X}$, $\mathcal{T}_{C \times D}$ is the table of prior information for $C$ cell types (classes) and $D$ markers (features), $\mathbbm{1}[\cdot]$ is an indicator function for partition membership, and $\{ \boldsymbol{\mu}_{k} \in \mathbb{R}^{D}, \boldsymbol{\Sigma}_{k} \in \mathbb{R}^{D \times D} \}$ are the parameters of the Normal distribution associated with each Mondrian partition $\mathcal{M}_{k}$.  Intuitively, the model can be thought of as a Gaussian mixture in which the mixture weights are determined by the MP.  If it were possible to integrate out the MP, the likelihood would be $\mathcal{L}(\mathbf{X}) = \prod_{i=1}^{N} \sum_{k} \pi_{\mathcal{M}_{k}} \ N(\mathbf{x}_{i}; \boldsymbol{\mu}_{k}, \boldsymbol{\Sigma}_{k})$ where $\pi_{\mathcal{M}_{k}}$ is the probability that $\mathbf{x}_{i}$ is contained in the $k$th partition.  As the number of partitions is random, we have a model similar in spirit to the Dirichlet process mixture model \cite{neal2000markov}, with the difference being the membership probabilities are determined by the MP's tree structure instead of the Dirichlet process' preferential attachment procedure (known as the \textit{Chinese restaurant process} \cite{ferguson1973bayesian}). 
\begin{figure}
\centering
\begin{subtable}{.40\textwidth}
\centering
\small
\begin{tabular}{l ccc }
\\
& \multicolumn{3}{c}{Markers (dimensions)} \\ 
Cell Types & CD4 & CD8  & CD3 \\
\hline 
Basophils  & $0$ & $-1$ & $-1$  \\ 
CD4 T cells  & $+1$  & $-1$ & $+1$  \\ 
CD8 T cells  & $-1$ & $+1$ & $+1$ \\
\end{tabular}
  \caption{Prior Information Table (Subsample)}
 % \label{fig:sub5p}
\end{subtable}
\begin{subtable}{.59\textwidth}
\centering
\small
\begin{tabular}{l l l }
\\
Marker Label Set & Dimension Prior & Cut Prior \\
\hline
 $\{ -1, 0, +1 \}$, $\{ -1, +1 \}$  & $p_{d} \propto \gamma_{0} \cdot |\mathcal{X}_{d}|$ & Beta($\phi_{0}$, $\phi_{0}$) \\
 $\{ -1, 0 \}$ & $p_{d} \propto \gamma_{1} \cdot |\mathcal{X}_{d}|$ & Beta($\phi_{1}$, $\phi_{0}$) \\
 $\{ 0, +1 \}$ & $p_{d} \propto \gamma_{1} \cdot |\mathcal{X}_{d}|$ & Beta($\phi_{0}$, $\phi_{1}$) \\
\end{tabular}
  \caption{Prior Distributions from Table Information}
 % \label{fig:sub5p}
\end{subtable}
\caption{\textit{Translating prior information to prior distributions}.  Subtable (a) shows the table of given information about how canonical cell types are expressed in each marker (dimension) \cite{lee2017automated}.  Subtable (b) demonstrates how we translate the prior information into prior distributions.  Given the current position in the Mondrian tree, the remaining labels in each dimension affect which dimension and how the cut is drawn.}
\label{priorInfo}
\end{figure}   

The prior information table $\mathcal{T}$ is used when drawing both the dimension $\hat{d}$ and the cut $\hat{c}$ at each step in the MP.  Starting with the former, dimensions are sampled according to $\hat{d} \sim \text{Multinoulli}(p_{d} \propto \gamma_{\mathcal{T}[d]}| \mathcal{X}_{d} |)$ where $| \mathcal{X}_{d}|$ is again the length and $\gamma_{\mathcal{T}[d]}$ returns a scalar value based on the set of labels observed in the corresponding column of the prior information table.  The scalars are set according to subtable (b) of Figure \ref{priorInfo}: dimensions with both high ($+1$) and low ($-1$) labels are upweighted by a factor of $\gamma_{0}$, making them more likely to be closer to the root of the tree.  This strategy is inspired by the use of information gain to build decision trees, which places the more discriminative features closer to the root.  Dimensions with only one informative label are upweighted by $\gamma_{1}$, which is set such that $\gamma_{1} << \gamma_{0}$.  As for drawing cuts, instead of drawing them uniformly over the dimension, we draw them according to three different Beta distributions, depending again on the column labels.  If the column labels for the current dimension contain both $+1$ and $-1$, we draw from Beta($\phi_{0}$, $\phi_{0}$).  Setting both Beta parameters to the same value concentrates the Beta's mass around $0.5$, which is appropriate since the cut should form both high and low regions.  If just $-1$ or $+1$ is present in the current table column, then Beta($\phi_{1}$, $\phi_{0}$) or Beta($\phi_{0}$, $\phi_{1}$) such that $\phi_{0} >  \phi_{1}$ are drawn from respectively.  After drawing from the Beta, we re-scale the cut point appropriately for the dimension, i.e. $\hat{c} = a_{d} + \hat{r}(b_{d}-a_{d})$ where $\hat{r} \sim \text{Beta}(\alpha_{\mathcal{T}[d]}, \beta_{\mathcal{T}[d]})$.  Lastly, we feed the appropraite sub-tables to the child Mondrian processes.  Using SQL notation, we perform \{\texttt{select * where $\mathcal{T}[\hat{d}] == -1$ or $\mathcal{T}[\hat{d}] == 0$}\} and feed the resulting subtable into the left child ($\mathcal{M}_{<c}$).  For the right child, we perform the same query with $+1$.  Thus upon each recursion, the table contains only the cell types that agree with the cut history.  Our model is related to Wang et al.\ (2015)'s method that incorporate meta-data into an MP by adjusting the size of the partitions \cite{wang2015metadata}.  Our proposed MP method also changes the multinoulli (dimension prior) based on prior information through re-scaling. One difference between our approach and that of Wang et al.\ is that we  use informative priors on the cut distributions, which  Wang et al.\ do not do.   

\textbf{Posterior Inference.}  Previous work \cite{roy2009mondrian, wang2015metadata} performed posterior inference via Markov chain Monte Carlo (MCMC) with proposals being drastic changes to the segmentation structure (translations, rotations, re-drawning, etc.\ of partitions).  We could not use this strategy since the priors for our MP are dependent on the tree's path, and re-calculating the prior for such global changes is not efficient.  Instead, we draw proposals from the prior and run MCMC on just the cut locations.  Local perturbations of the cuts are proposed, and these are run until convergence.  %There is a chance that we drew a poor tree (i.e. cut ordering) and cannot reach the areas of high posterior probability because the sample can be edited only locally.  Hence, before accepting a chain's samples, we perform a `burn-in' test in which we compare their log likelihood against a k-means maximum likelihood solution, accepting if the samples' likelihood is within a threshold.

\textbf{Classifying Cells into Types.}  Our model is completely unsupervised at the cell level, taking no account of cell-level labels during training.  We perform cell type classification by using the information table combined with the MP posterior distribution.  We find the MP partitions that obey the table constraints for each type.  For example, CD4 T cells are assigned to the partition on the high side of the cut on CD4, low side of CD8, high side of CD3, etc.  In the rare case that two or more cell types are assigned to a partition, we draw one of the remaining labels at random.  Final classification accuracy is determined by voting across all posterior samples.

\begin{figure}
\centering
\begin{subfigure}{.28\textwidth}
\centering
\includegraphics[width=.95\linewidth]{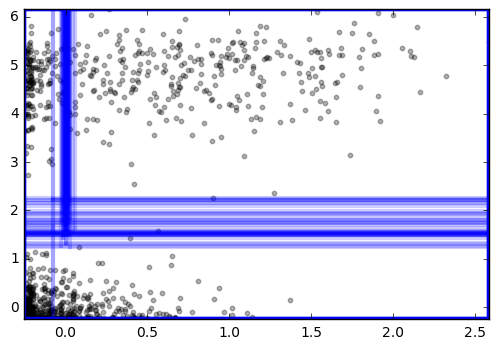}
\caption{\centering{100 Posterior Samples AML: CD4 vs CD3}}
\label{fig:posterior_AML}
\end{subfigure}
\hfil
\begin{subfigure}{.28\textwidth}
\centering
\includegraphics[width=.99\linewidth]{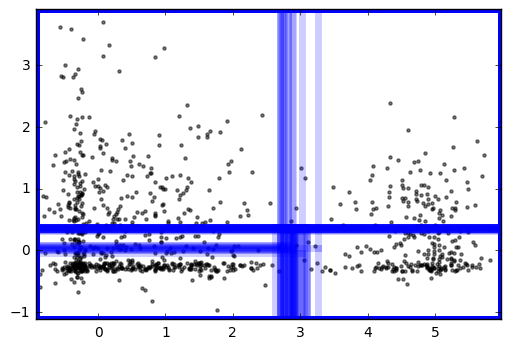}
\caption{\centering{100 Posterior Samples BMMC: CD4 vs CD8}}
\label{fig:posterior_BMMC}
\end{subfigure}
\hfil
\begin{subtable}{.30\textwidth}
\centering
\small
\begin{tabular}{l cc}
\\
 & AML & BMMC   \\
\hline \hline
ACDC \cite{lee2017automated} & $98.3\%$ & $92.9\%$\text{*} \\
MP-GMM  & $92.2\%$ & $93.3\%$  \\ 
\hline 
GMM  & $86.1\%$ & $84.1\%$ \\
MP-Prior  & $61.5\%$  & $85.6\%$  \\ 
 \\
\end{tabular}
  \caption{\centering{Accuracy on Cell Type Classification}}
  \label{fig:accuracy}
\end{subtable}
\caption{\textit{Results on AML and BMMC Datasets}.  Subfigures (a) and (b) show 100 posterior samples for two markers on each dataset.  Subtable (c) shows classification accuracy on identifying canonical cell types.  * indicates that the Lee et al.\ (2015) BMMC result includes an extra `unknown' class and thus is \emph{not} directly comparable to our reported accuracy.}
\label{Results}
\end{figure}   

\vspace{-0.1in}
\section{Experimental Results}
\vspace{-0.1in}
We evaluated our method on the same human samples and prior information tables used in \cite{lee2017automated}.  The samples are from two public benchmark datasets: acute myeloid leukemia (AML) \cite{levine2015data} and bone marrow mononuclear cells (BMMC) \cite{bendall2011single}. 
%Both of AML and BMMC datasets were collected from healthy human bone marrow. 
The AML dataset consists of 32 markers and has been manually gated into 14 cell types. The BMMC sample has 13 markers and 19 cell types.  
%We transformed all marker values using $sinh^{-1}((x-1)/5)$ as in \cite{lee2017automated}. 
In all experiments, we set the informative priors as follows: $\gamma_{0}=100$, $\gamma_{1}=1$, $\phi_{0}=5$, $\phi_{1}=2$.  For inference, we drew 50 samples from the prior and ran 2000 iterations of MCMC on each.

\textbf{Visualizing Posterior Samples.}  For our first experimental result, we demonstrate the interpretability of the MP's posterior.  Subfigures (a) and (b) of Figure \ref{Results} show posterior samples for the AML and BMMC datasets.  Each blue line represents a sampled cut and each black dot represents a cell for the two plotted markers (CD4 vs CD3 for AML, CD4 vs CD8 for BMMC).  One can imagine showing a technician successive 2D plots involving the dimension cut at each step in the tree.  Furthermore, these samples could be integrated into an interactive user interface for gating, to show the model's suggestions for cuts, including its uncertainty about the cut based on the dispersion of the samples.  Figure \ref{Histogram-Tree} in the Appendix shows a MAP estimate of the full tree.  Red lines are sampled cuts, arrows denote a cells path through the tree depending on which side of the cut it falls, and black boxes denote type classifications.       

\textbf{Classification Accuracy.}
Figure \ref{Results} (c) reports cell type classfication accuracy (relative to manual gating) of our method (MP-GMM) against ACDC \cite{lee2017automated} and two baselines: a Gaussian mixture model (GMM) and samples from the MP prior (MP-Prior).  We see that our model outperforms a mixture model baseline with no prior information (GMM) and an MP model with no dependence on the data (MP-Prior).  Relative to the ACDC algorithm, our model's  performance is close  but slightly worse. However, our model does not require cell-level labels (which ACDC does) and it is arguably more interpretable to flow cytometry experts than ACDC since it more directly mimics the manual gating process. 

\vspace{-0.1in}
\section{Conclusions and Future Work}
\vspace{-0.1in}
We have presented preliminary work on using Mondrian processes (MPs) for classification of cells into canonical types.  While our method achieves competitive but not superior accuracy in comparison to other state-of-the-art methods \cite{lee2017automated}, we believe the simplicity of our model as well as its intuitive tree structure can facilitate the method's integration into clinical practice.  Furthermore, as shown in the experimental section, visualization of posterior samples provides an intuitive window into the model's behavior.  

%One can imagine these being displayed within a gating user-interface, offering the technician a guide as to what the model would do, including a notion of certainty based on how the cuts concentrate.  
There are several directions for future work on both practical and theoretical aspects.  In terms of application to flow cytometry, further tuning of the method would likely be needed to achieve state of the art accuracy.  This improvement can be done through more hyperparameter exploration (we did relatively little) as well as minor changes to the model.  For example, we could change the mixture distribution to something other than Normal, or  add supervison during training as Lee et al. (2015) do.  Extending the model to perform analysis on multiple individuals at once is another attractive extension, and this motivates theoretical work on MPs.  To the best of our knowledge, a hierarchcial formulation of the MP has not been explored; such a model could model each individual with an MP while sharing statistical strength across trees. 

\section*{Acknowledgments}
The authors thank Dr.\ Yu Qian and Dr.\ Richard Scheuermann of the J.\ Craig Venter Institute for their valuable assistance. The research described in this paper was supported in part by the National Center For Advancing Translational Sciences of the National Institutes of Health under award number U01TR001801 and by the National Science Foundation under award number IIS-1320527. The content is solely the responsibility of the authors and does not necessarily represent the official views of the National Institutes of Health or the National Science Foundation.

%\newpage
%\clearpage
\bibliography{cytometry}
\bibliographystyle{plain}

\newpage
\appendix
\section*{Appendix}
%\subsection*{Mondrian Tree}
\begin{figure}[h!]
\centering
\includegraphics[width=1.3\linewidth, angle = -90]{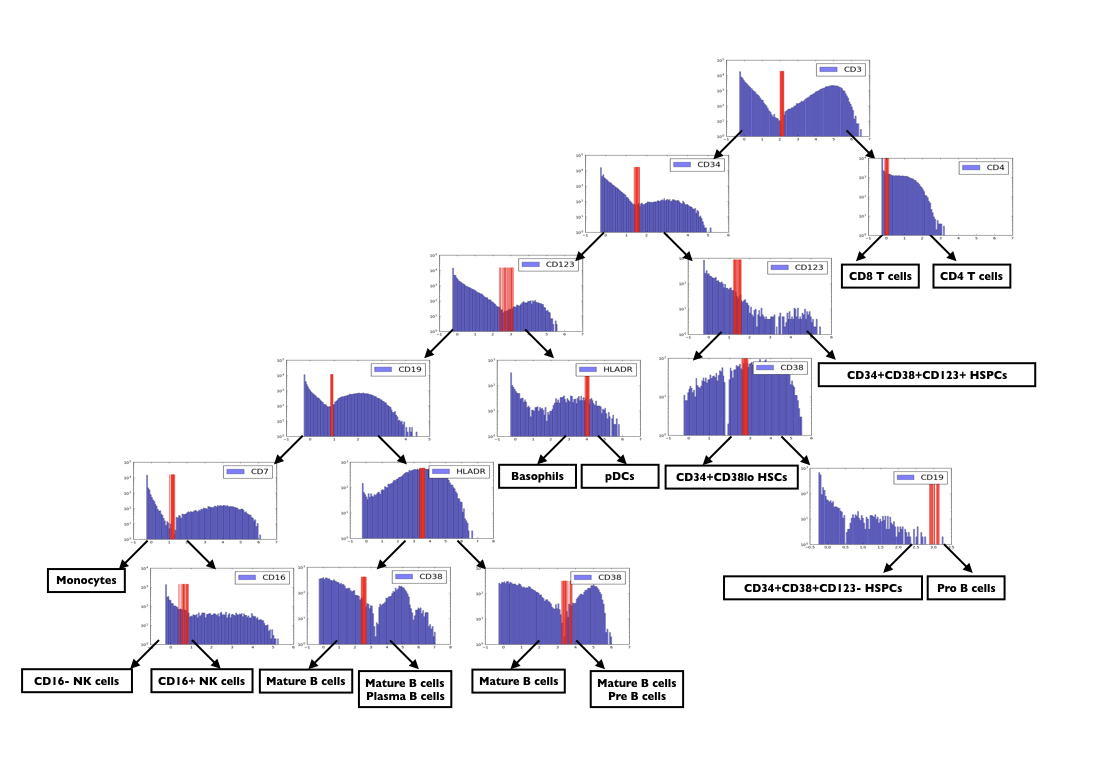}
\caption{\textit{Visualization of Mondrian Tree}.  The figure above shows the tree structure of the posterior sample with highest likelihood (MAP estimate) on the AML dataset.  Red lines denote sampled cuts, and arrows denote the path taken by cells that fall on the left or right side of the cut.  The black rectangles denote cell type classifications.}
\label{Histogram-Tree}
\end{figure}   

\end{document}